\def\year{2021}\relax
\documentclass[letterpaper]{article} 

\newcommand{\final}{0}
\usepackage{aaai21}  
\usepackage{times}  
\usepackage{helvet} 
\usepackage{courier}  
\usepackage[hyphens]{url}  
\usepackage{graphicx} 
\usepackage{natbib}  
\usepackage{caption} 
\usepackage{booktabs}
\usepackage{amsmath}
\usepackage{amssymb}
\usepackage{epsfig}
\usepackage{color}
\usepackage{microtype}
\usepackage{algorithm}
\usepackage{algorithmic}
\usepackage{ifthen}
\usepackage[switch]{lineno}  %
\usepackage{engord}
\usepackage{ifthen}
\definecolor{WeimingColor}{rgb}{0,0,0.8} 
\definecolor{YingyingColor}{rgb}{0.0,0.8,0.8} 
\definecolor{FanColor}{rgb}{0.8,0,0.8}

\newcommand{\weiming}[1]{{\color{WeimingColor} [Weiming: #1]}}

\newcommand{\fan}[1]{{\color{FanColor}[Fan: #1]}}

\newcommand{\warning}[1]{{\it\color{red} #1}}
\newcommand{\toremove}[1]{{\it\color{red} (To remove) #1}}
\newcommand{\note}[1]{{\it\color{blue} #1}}
\newcommand{\nothing}[1]{}

\ifthenelse{\equal{\final}{1}}
{
\renewcommand{\weiming}[1]{}
\renewcommand{\fan}[1]{}
\renewcommand{\yingying}[1]{}

\renewcommand{\warning}[1]{}
\renewcommand{\toremove}[1]{}
\renewcommand{\note}[1]{}
\renewcommand{\nothing}[1]{}
}
{}

\title{Arbitrary Video Style Transfer via Multi-Channel Correlation}

\author{
Yingying Deng\textsuperscript{\rm 1,2,4},
\hspace{20pt}
Fan Tang\textsuperscript{\rm 3}$^*$,
\hspace{20pt}
Weiming Dong\textsuperscript{\rm 1,2,4}\thanks{Co-corresponding authors},
\\
Haibin Huang\textsuperscript{\rm 5},
\hspace{12pt}
Chongyang Ma\textsuperscript{\rm 5},
\hspace{12pt}
Changsheng Xu\textsuperscript{\rm 1,2,4}
\\
}
\affiliations{
\textsuperscript{\rm 1} School of Artificial Intelligence, University of Chinese Academy of Sciences,
\textsuperscript{\rm 2} NLPR, Institute of Automation, Chinese Academy of Sciences,
\textsuperscript{\rm 3} School of Artificial Intelligence, Jilin University,
\textsuperscript{\rm 4} CASIA-LLvision Joint Lab,
\textsuperscript{\rm 5} Kuaishou Technology
\\
\{dengyingying2017, weiming.dong, changsheng.xu\}@ia.ac.cn, tangfan@jlu.edu.cn, \\
\{huanghaibin03, chongyangma\}@kuaishou.com
}

\begin{document}
%
\maketitle

\begin{abstract}
Video style transfer is attracting increasing attention from the artificial intelligence community because of its numerous applications, such as augmented reality and animation production.
Relative to traditional image style transfer, video style transfer presents new challenges, including how to effectively generate satisfactory stylized results for any specified style while maintaining temporal coherence across frames. 
Towards this end, we propose a Multi-Channel Correlation network (MCCNet), which can be trained to fuse exemplar style features and input content features for efficient style transfer while naturally maintaining the coherence of input videos to output videos. 
Specifically, MCCNet works directly on the feature space of style and content domain where it learns to rearrange and fuse style features on the basis of their similarity to content features. 
The outputs generated by MCC are features containing the desired style patterns that can further be decoded into images with vivid style textures. 
Moreover, MCCNet is also designed to explicitly align the features to input and thereby ensure that the outputs maintain the content structures and the temporal continuity. 
To further improve the performance of MCCNet under complex light conditions, we also introduce illumination loss during training. 
Qualitative and quantitative evaluations demonstrate that MCCNet performs well in arbitrary video and image style transfer tasks. Code is available at \url{https://github.com/diyiiyiii/MCCNet}.

\end{abstract}

\section{Introduction}

Style transfer is a significant topic in the industrial community and research area of artificial intelligence. 
Given a content image and an art painting, a desired style transfer method can render the content image into the artistic style referenced by the art painting.
Traditional style transfer methods based on stroke rendering, image analogy, or image filtering~\citep{efros:2001:image,bruckner:2007:style,strothotte:2002:non} only use low-level features for texture transfer~\citep{gatys:2016:image,Doyle:2019:APM}.
Recently, deep convolutional neural networks (CNNs) have been widely studied for artistic image generation and translation~\citep{gatys:2016:image,johnson:2016:perceptual,zhu:2017:unpaired,Huang:2017:Arbitrary,huang:2018:multimodal,jing:2020:dynamic}. 

\begin{figure}
\setlength{\abovecaptionskip}{2mm}
\centering
\includegraphics[width=\linewidth]{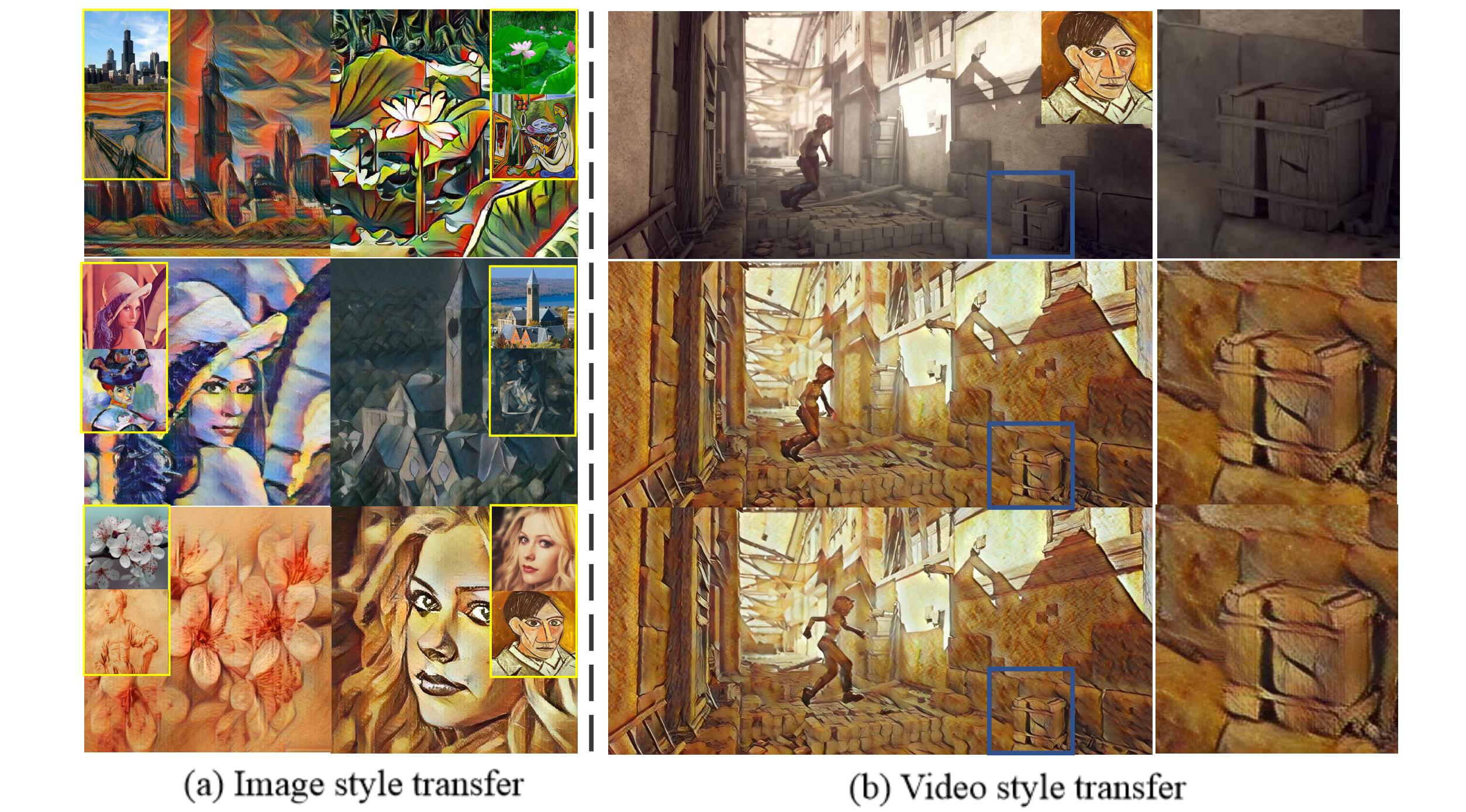}
\caption{The proposed method can be used for stable video style transfer with high-quality stylization effect: (a) image style transfer results using images in different domains; (b) video style transfer results using different frames from \textit{Sintel}. The stylization results of the same object in different frames are the same. Hence, the rendered effects for the same content are stable between different frames.}
\label{fig:caption}
\end{figure}

Although existing methods can generate satisfactory results for still images, they lead to flickering effects between adjacent frames when applied to videos directly~\citep{ruder:2016:artistic,chen:2017:coherent}.
Video style transfer is a challenging task that needs to not only generate good stylization effects per frame but also consider the continuity between adjacent frames of the video.
\citet{ruder:2016:artistic} added temporal consistency loss on the basis of the approach proposed in \citet{gatys:2016:image} to maintain a smooth transition between video frames.
\citet{chen:2017:coherent} proposed an end-to-end framework for online video style transfer through a real-time optical flow and mask estimation network. 
\citet{gao:2020:fast} proposed a multistyle video transfer model, which estimates the light flow and introduces temporal constraint.
However, these methods highly depend on the accuracy of optical flow calculation, and the introduction of temporal constraint loss reduces the stylization quality of individual frames.
Moreover, adding optical flow to constrain the coherence of stylized videos makes the network difficult to train when applied to an arbitrary style transfer model.  

In the current work, we revisit the basic operations in state-of-the-art image stylization approaches and propose a frame-based \textit{Multi-Channel Correlation} network (MCCNet) for temporally coherent video style transfer that does not involve the calculation of optical flow.
Our network adaptively rearranges style representations on the basis of content representations by considering the multi-channel correlation of these representations. Through this approach, MCCNet is able to make style patterns suitable for content structures.
By further merging the rearranged style and content representations, we generate features that can be decoded into stylized results with clear content structures and vivid style patterns.
MCCNet aligns the generated features to content features, and thus, slight changes in adjacent frames will not cause flickering in the stylized video.
Furthermore, the illumination variation among consecutive frames influences the stability of video style transfer. 
Thus, we add random Gaussian noise to the content image to simulate illumination varieties and propose an illumination loss to make the model stable and avoid flickering. 
As shown in Figure~\ref{fig:caption}, our method is suitable for stable video style transfer, and it can generate single-image style transfer results with well-preserved content structures and vivid style patterns.
In summary, our main contributions are as follows:
\begin{itemize}

\item We propose MCCNet for framed-based video style transfer by aligning cross-domain features with input videos to render coherent results.

\item We calculate the multi-channel correlation across content and style features to generate stylized images with clear content structures and vivid style patterns.

\item 
We propose an illumination loss to make the style transfer process increasingly stable so that our model can be flexibly applied to videos with complex light conditions.
\end{itemize}

\section{Related Work}

\paragraph{Image style transfer.}
Image style transfer has been widely studied in recent years. Essentially, it enables the generation of artistic paintings without the expertise of a professional painter.
\citet{gatys:2016:image} found that the inner products of the feature maps in CNNs can be used to represent style and proposed a neural style transfer (NST) method through continuous optimization iterations. 
However, the optimization process is time consuming and cannot be widely used.
\citet{johnson:2016:perceptual} put forward a real-time style transfer method to dispose of a specific style in one model.
\citet{dumoulin:2016:learned} proposed conditional instance normalization (CIN), which allows the learning of multiple styles in one model by reducing a style image into a point in the embedded space.
A number of methods achieve arbitrary style transfer by aligning the second-order statistics of style and content images~\citep{Huang:2017:Arbitrary,li:2017:universal,wang:2020:diversified}.
\citet{Huang:2017:Arbitrary} proposed an arbitrary style transfer method by adopting adaptive instance normalization (AdaIN), which normalizes content features using the mean and variance of style features.
\citet{li:2017:universal} used whitening and coloring transformation (WCT) to render content images with style patterns.
\citet{wang:2020:diversified} adopted deep feature perturbation (DFP) in a WCT-based model to achieve diversified arbitrary style transfer.
However, these holistic transformations lead to unsatisfactory results.
\citet{park:2019:arbitrary} proposed a style-attention network (SANet) to obtain abundant style patterns in generated results but failed to maintain distinct content structures.
\citet{yao:2019:attention} proposed an attention-aware multi-stroke style transfer (AAMS) model by adopting self-attention to a style swap-based image transfer method, which highly relies on the accuracy of the attention map used.
\citet{deng:2020:arbitrary} proposed a multi-adaptation style transfer (MAST) method to disentangle content and style features and combine them adaptively. 
However, some results are rendered with uncontrollable style patterns.

In the current work, we propose an arbitrary style transfer approach, which can be applied to video transfer with better stylized results than other state-of-the-art methods. 

\paragraph{Video style transfer.}
Most video style transfer methods rely on existing image style transfer methods~\citep{ruder:2016:artistic,chen:2017:coherent,gao:2020:fast}. 
\citet{ruder:2016:artistic} built on NST and added a temporal constraint to avoid flickering. 
However, the optimization-based method is inefficient for video style transfer. 
\citet{chen:2017:coherent} proposed a feed-forward network for fast video style transfer by incorporating temporal information.
\citet{chen:2020:optical} distilled knowledge from the video style transfer network with optical flow to a student network to avoid optical flow estimation in the test stage.
\citet{gao:2020:fast} adopted CIN for multi-style video transfer, and the approach incorporated one FlowNet and two ConvLSTM modules to estimate light flow and introduce a temporal constraint.
The temporal constraint aforementioned is achieved by calculating optical flow, and the accuracy of optical flow estimation affects the coherence of the stylized video.
Moreover, the diversity of styles is limited because of the used basic image style transfer methods used. 
\citet{wang:2020:consistent} proposed a novel interpretation of temporal consistency without optical flow estimation for efficient zero-shot style transfer.
\citet{li:2019:learning} learned a transformation matrix for arbitrary style transfer. 
They found that the normalized affinity for generated features are the same as that for content features, and is thus suitable for frame-based video style transfer.
However, the style patterns in their stylized results are not clearly noticeable.

To avoid using optical flow estimation while maintaining video continuity, we aim to design an alignment transform operation to achieve stable video style transfer with vivid style patterns.

\section{Methodology}

\begin{figure*}
\setlength{\abovecaptionskip}{2mm}
\centering
\includegraphics[width=\linewidth]{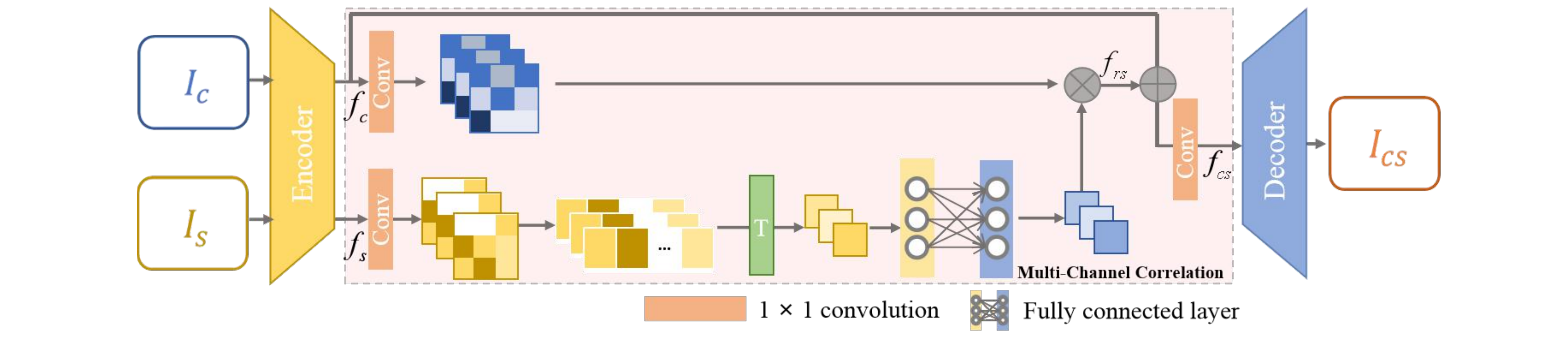}
\caption{Overall structure of MCCNet. The green block represents the operation of the feature vector multiplied by the transpose of the vector. $\otimes$ and $\oplus$ represent the matrix multiplication and addition. $I_c$, $I_s$ and $I_{cs}$ refer to the content, style and generated images, respectively. 
} 
\label{fig:network}
\end{figure*}

\begin{figure}
\setlength{\abovecaptionskip}{2mm}
\centering
\includegraphics[width=\linewidth]{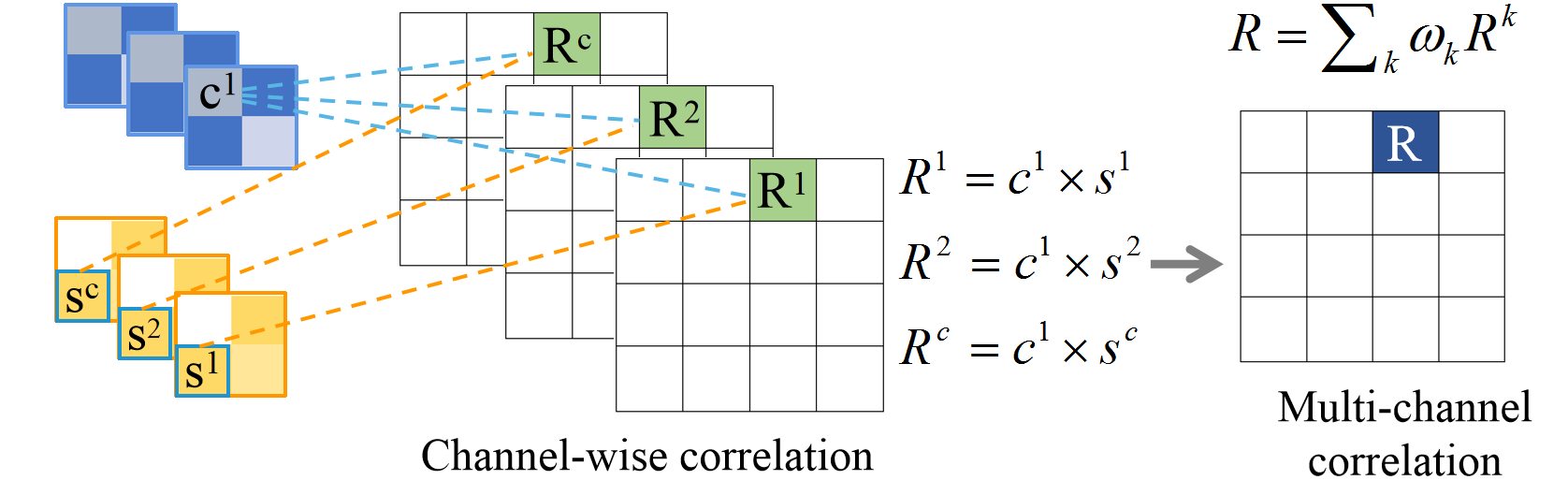}
\caption{Correlation matrix CO. $R$ is the element of CO.}
\label{fig:technical}
\end{figure}

As shown in Figure~\ref{fig:network}, the proposed MCCNet adopts an encoder-decoder architecture. 
Given a content image $I_c$ and a style image $I_s$, we can obtain corresponding feature maps $f_c = E(I_c)$ and $f_s = E(I_s)$ through the encoder. 
Through MCC calculation, we generate $f_{cs}$ that can be decoded into stylized image $I_{cs}$.
 
We first formulate and analyze the proposed multi-channel correlation in Sections~\ref{sec:mcc_formu} and~\ref{sec:mcc_anay} and then introduce the configuration of MCCNet involving feature alignment and fusion in Section~\ref{sec:mcc_train}.

\subsection{Multi-Channel Correlation}
\label{sec:mcc_formu}
 
Cross-domain feature correlation has been studied for image stylization~\cite{park:2019:arbitrary,deng:2020:arbitrary}.
It fuses the multiple content and style features by using several adaption/attention modules without considering the inter-channel relationship of the content features.
In SANet~\cite{park:2019:arbitrary}, the generated features can be formulated as
\begin{equation}
f_cs=F(f_c,f_s) \propto exp(f(f_c),g(f_s))h(f_s). 
\end{equation}
Slight changes in input content features can lead to a large-scale variation in output features. 
MAST~\cite{deng:2020:arbitrary} has the same issue.
Thus the coherence of input content features cannot be migrated to generated features, and the stylized videos will present flickering artifacts.

In this work, we propose the multi-channel correlation for frame-based video stylization, as illustrated in Figure~\ref{fig:technical}.
For channel $i$, the content and style features are $f_c^i \in \mathbb{R}^{ H \times W}$ and $f_s^i \in \mathbb{R}^{H \times W}$, respectively.
We reshape them to $f_c^i \in \mathbb{R}^{ 1 \times N}, f_c^i = [c_1, c_2,\cdots,c_N]$ and $f_s^i \in \mathbb{R}^{1 \times N}, f_s^i = [s_1, s_2,\cdots,s_N]$, where $N = H \times W $.
The channel-wise correlation matrix between content and style is calculated by
\begin{equation}
CO^i = f_{c}^{i T} \otimes f_s^i.
\end{equation}
Then, we rearrange the style features by
\begin{equation}
f_{rs}^i = f_{s}^i \otimes CO^{i T} 
=\|f_{s}^i\|_2 f_c^i,
\label{fun:2}
\end{equation}
where $\|f_{s}^i\|_2 = \sum_{j=1}^{N} s_j^2$.
Finally, the channel-wise generated features are
\begin{equation}
f_{cs}^i = f_c^i +f_{rs}^i   =(1+ \|f_{s}^i\|_2)f_c^i.
\end{equation}


However, the multi-channel correlation in style features is also important to represent style patterns (e.g., texture and color).
We calculate the correlation between each content channel and every style channel. 
The $i$-th channel for generated features can be rewritten as
\begin{equation}
f_{cs}^i =   f_c^i +f_{rs}^i =(1+ \sum_{k=1}^{C}w_k \|f_{s}^k\|_2)f_c^i,
\label{fun:4}
\end{equation}
where $C$ is the number of channels and $w_k$ represents the weights of the $k$-th style channel.
Finally, the generated features $f_{cs}$ can be obtained by
\begin{equation}
f_{cs} =K f_c,
\label{fun:5}
\end{equation}
where $K$ is the style information learned by training.
\begin{figure}
\setlength{\abovecaptionskip}{2mm}
\centering
\includegraphics[width=\linewidth]{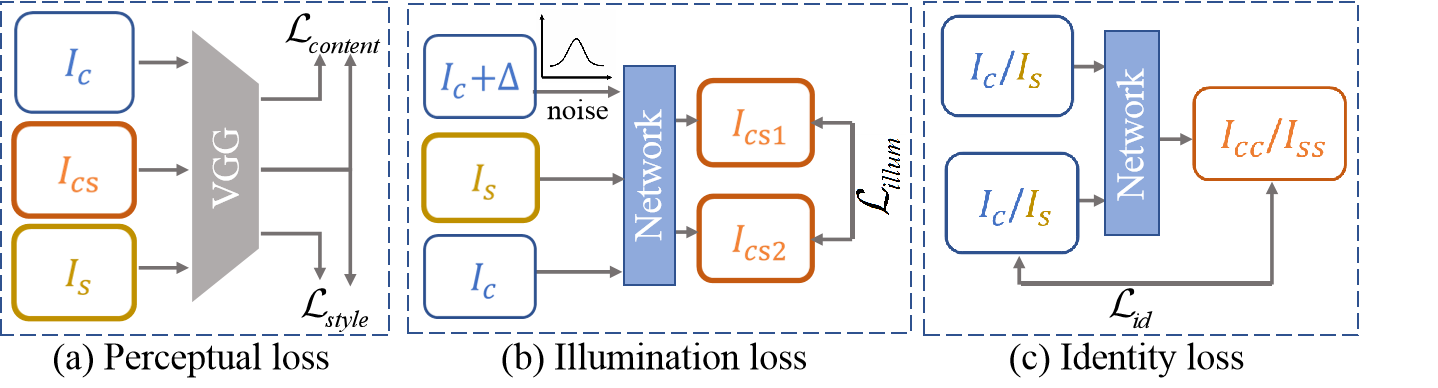}
\caption{Loss function.}
\label{fig:loss}
\end{figure}

From Equation~(\ref{fun:5}), we can conclude that MCCNet can help to generate stylized features that are strictly aligned to content features.
Therefore, the coherence of input videos is naturally maintained to output videos and slight changes (e.g., objects motions) in adjacent frames cannot lead to violent changes in the stylized video.
Moreover, through the correlation calculation between content and style representations, the style patterns are adjusted according to content distribution. The adjustment helps generate stylized results with stable content structures and appropriate style patterns.

\subsection{Coherence Analysis}
\label{sec:mcc_anay}
A stable style transfer requires that the output stylized video to be as coherent as the input video.
As described in \citet{wang:2020:consistent}, a coherent video should satisfy the following constraint:
\begin{equation}
\lVert X^m -W_{X^n \rightarrow X^m}X^n \lVert < \delta,
\label{equ:consis}
\end{equation}
where $X^m, X^n$ are the $m$-th and $n$-th frames, respectively; and $W_{X^n \rightarrow X^m} $ is the warping matrix from $X^m$ to $X^n$.
$\delta$ is a minimum so that humans are not aware of the minor flickering artifacts.
When the input video is coherent, we obtain the content features $f_c$ of each frame. And the content features of the $m$-th and $n$-th frame satisfy
\begin{equation}
\lVert f_c^m -W f_c^n \lVert < \delta.
\end{equation}
For corresponding output features $f_{cs}^m$ and $f_{cs}^n$:
\begin{equation}
\begin{split}
\lVert f_{cs}^m -W f_{cs}^n \lVert &= \lVert K f_{c}^m -W  K f_{c}^n \lVert \\
&=|K| \cdot \lVert  f_{c}^m -W f_{c}^n \lVert < |K|  \cdot \delta.
\end{split}
\end{equation}
Then, the output video also satisfies:
\begin{equation}
\lVert f_{cs}^m -W f_{cs}^n \lVert < \gamma,
\end{equation}
where $\gamma$ is a minimum.
Therefore, our network can migrate the coherence of the input video frame features to the stylized frame features without additional temporal constraints.
We further demonstrate that the coherence of stylized frame features can be well-transited to the generated video despite the convolutional operation of the decoder in Section~\ref{sec:ablation}.
Such observations prove that the proposed MCCNet is suitable for video style transfer tasks.

\subsection{Network Structure and Training}

\label{sec:mcc_train}
In this section, we introduce how to involve MCC in the encoder-decoder based stylization framework (Figure~\ref{fig:network}). 
Given $f_c$ and $f_s$, we first normalize them and feed them to the $1 \times 1$ convolution layer. 
We then stretch  $f_s \in \mathbb{R}^{C \times H \times W}$ to  $f_s \in \mathbb{R}^{C \times 1 \times N}$, and $f_s f_s^T = [\|f_{s}^1\|_2,\|f_{s}^2\|_2,...,\|f_{s}^C\|_2 ]$ is obtained through covariance matrix calculation. 
Next, we add a fully connected layer to weigh the different channels in $f_s f_s^T$. 
Through matrix multiplication, we weigh each content channel with a compound of different channels in $f_s f_s^T$ to obtain $f_{rs}$.
Then, we add $f_c$ and $f_{rs}$ to obtain $f_{cs}$ defined in Equation~\ref{fun:4}. 
Finally, we  feed $f_{cs}$ to a $1 \times 1$ convolutional layer and decode it to obtain the generated image $I_{cs}$.

As shown in Figure~\ref{fig:loss}, our network is trained by minimizing the loss function defined as
\begin{equation}
\begin{split}
\mathcal{L} &= \lambda_{content}\mathcal{L}_{content}  + \lambda_{style}   \mathcal{L}_{style}  \\
&+\lambda_{id}\mathcal{L}_{id} + \lambda_{illum}\mathcal{L}_{illum}.
\end{split}
\end{equation}
The total loss function includes perceptual loss $\mathcal{L}_{content}$ and $\mathcal{L}_{style}$, identity loss $\mathcal{L}_{id}$ and illumination loss $\mathcal{L}_{illum}$ in the training procedure.
The weights $\lambda_{content}$, $\lambda_{style}$, $\lambda_{id}$, and $\lambda_{illum}$ are set to $4$, $15$, $70$, and $3,000$ to eliminate the impact of magnitude differences.

\paragraph{Perceptual loss.}
We use a pretrained VGG19 to extract content and style feature maps and compute the content and style perceptual loss similar to AdaIN~\cite{Huang:2017:Arbitrary}.
In our model, we use layer $conv4\_1$ to calculate the content perceptual loss and layers $conv1\_1$, $conv2\_1$, $conv3\_1$, and $conv4\_1$ to calculate the style perceptual loss.
The content perceptual loss $\mathcal{L}_{content}$ is used to minimize the content differences between generated images and content images, where
\begin{equation}
\mathcal{L}_{content} = \lVert \phi_i(I_{cs}) - \phi_i(I_{c})  \lVert_2.
\end{equation}
The style perceptual loss $\mathcal{L}_{style}$ is used to minimize the style differences between generated images and style images:
\begin{equation}
\begin{split}
\mathcal{L}_{style} = \sum_{i=1}^{L} \mathcal{L}^i_{style},
\end{split}
\end{equation}
\begin{equation}
\begin{split}
\mathcal{L}^i_{style} &= \lVert \mu(\phi_i(I_{cs})) - \mu(\phi_i(I_{s}))  \lVert_2 \\
&+ \lVert \sigma(\phi_i(I_{cs})) - \sigma(\phi_i(I_{s}))  \lVert_2,
\end{split}
\end{equation}
where $\phi_i(\cdot)$ denotes the features extracted from the $i$-th layer in a pretrained VGG19, $\mu(\cdot)$ denotes the mean of features, and $\sigma (\cdot)$ denotes the variance of features.

\paragraph{Identity loss.}
We adopt the identity loss to constrain the mapping relation between style features and content features, and help our model to maintain the content structure without losing the richness of the style patterns.
The identity loss $\mathcal{L}_{id}$ is defined as:
\begin{equation}
\mathcal{L}_{id} = \lVert I_{cc} - I_{c}  \lVert_2 + \lVert I_{ss} - I_{s}  \lVert_2,
\end{equation}
where $I_{cc}$ denotes the generated results using a common natural image as content image and style image and $I_{ss}$ denotes the generated results using a common painting as content image and style image.
\begin{figure*}
\setlength{\abovecaptionskip}{2mm}
\centering
\includegraphics[width=\linewidth]{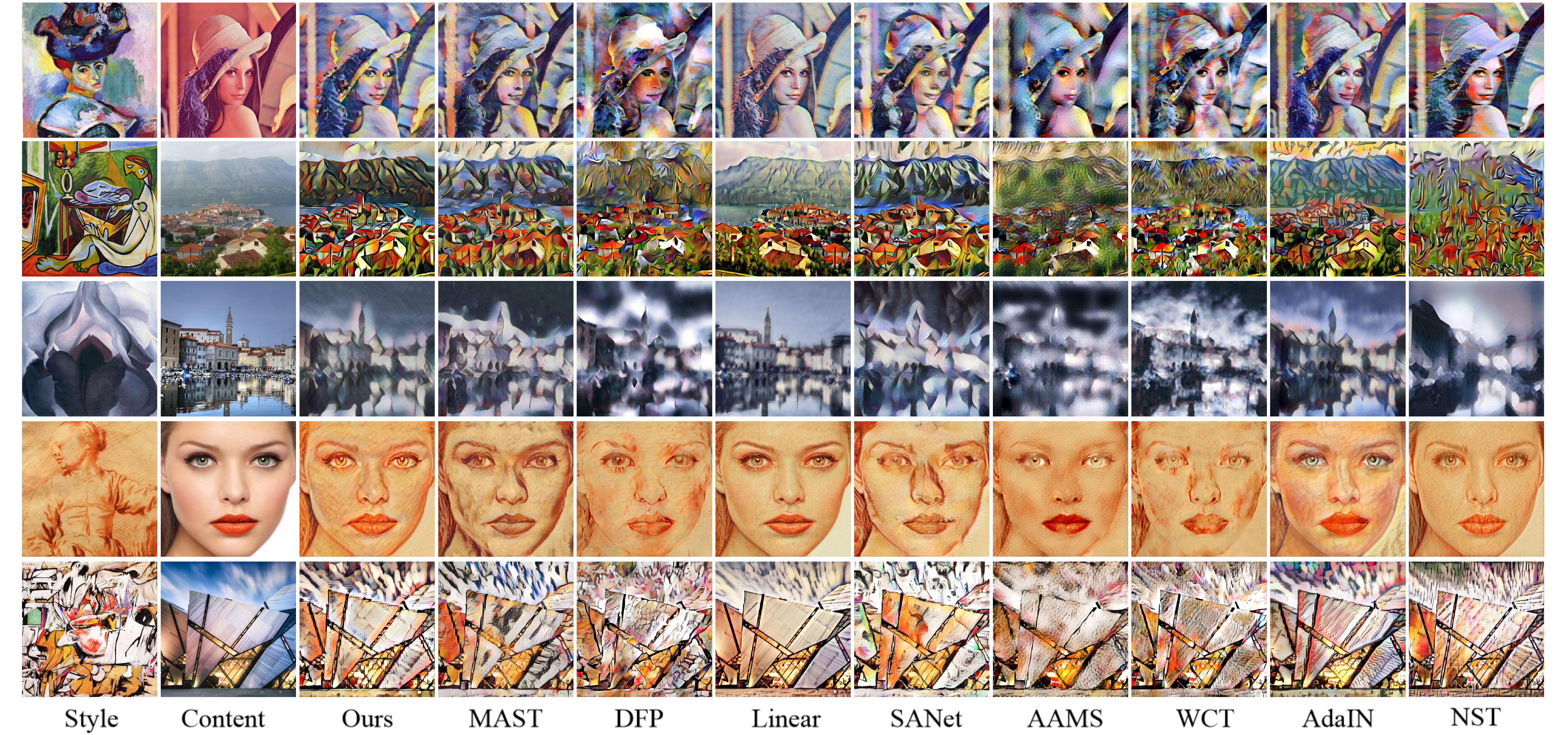}

\caption{Comparison of image style transfer results. The first column shows style images, the second column shows content images. The remaining columns are stylized results of different methods.
}
\label{fig:artist}
\end{figure*}

\paragraph{Illumination loss.}
For a video sequence, the illumination may change slightly that is difficult to be discovered by humans. The illumination variation in video frames could influence the final transfer results and result in flicking. 
Therefore, we add random Gaussian noise to the content images to simulate light. The illumination loss is formulated as
\begin{equation}
\mathcal{L}_{Illum} =\lVert G(I_c,I_s) - G(I_c+ \Delta,I_s) \lVert_2,
\end{equation}
where $G(\cdot)$ is our generation function, $\Delta \sim \mathcal{N}(0, \sigma^2 I)$.
With illumination loss, our method can be robust to complex light conditions in input videos.

\section{Experiments}
\begin{table}[t]
\centering

\begin{tabular}{cccccc}
\toprule
Image size&256& 512& 1024 \\
\midrule
Ours&0.013& 0.015& 0.019 \\

MAST&0.030&0.096& 0.506 \\

CompoundVST&0.049&0.098&0.285 \\

DFP&0.563&0.724& 1.260 \\

Linear&0.010&0.013& 0.022 \\

SANet&0.015& 0.019& 0.021 \\

AAMS&2.074& 2.173& 2.456 \\

AdaIN&0.007&0.008& 0.009 \\

WCT&0.451& 0.579& 1.008 \\
 
NST&19.528& 37.211& 106.372 \\
\bottomrule
\end{tabular}
\caption{Inference time of different methods.}
\label{tab:speed}
\end{table}
Typical video stylization methods use temporal constraint and optical flow to avoid flickering in generated videos. 
Our method focuses on promoting the stability of the transform operation in the arbitrary style transfer model on the basis of single frame.
Thus, the following frame-based SOTA stylization methods are selected for comparison: MAST~\citep{deng:2020:arbitrary}, CompoundVST~\citep{wang:2020:consistent}, DFP~\citep{wang:2020:diversified}, Linear~\citep{li:2019:learning}, SANet~\citep{park:2019:arbitrary}, AAMS~\citep{yao:2019:attention}, WCT~\citep{li:2017:universal}, AdaIN~\citep{Huang:2017:Arbitrary}, and NST~\citep{gatys:2016:image}.

In this section, we start from the training details of the proposed approach and then move on to the evaluation of image (frame) stylization and the analysis of rendered videos.

\subsection{Implementation Details and Statistics}
We use MS-COCO~\citep{lin:2014:coco} and WikiArt~\citep{phillips:2011:wiki} as the content and style image datasets for network training.
At the training stage, the images are randomly cropped to $256 \times 256$ pixels.
At the inference stage, images in arbitrary size are acceptable.
The encoder is a pretrained VGG19 network.
The decoder is a mirror version of the encoder, except for the parameters that need to be trained.
The training batch size is $8$, and the whole model is trained through $160,000$ steps.

\begin{figure}
\setlength{\abovecaptionskip}{2mm}
\centering
\includegraphics[width=\linewidth]{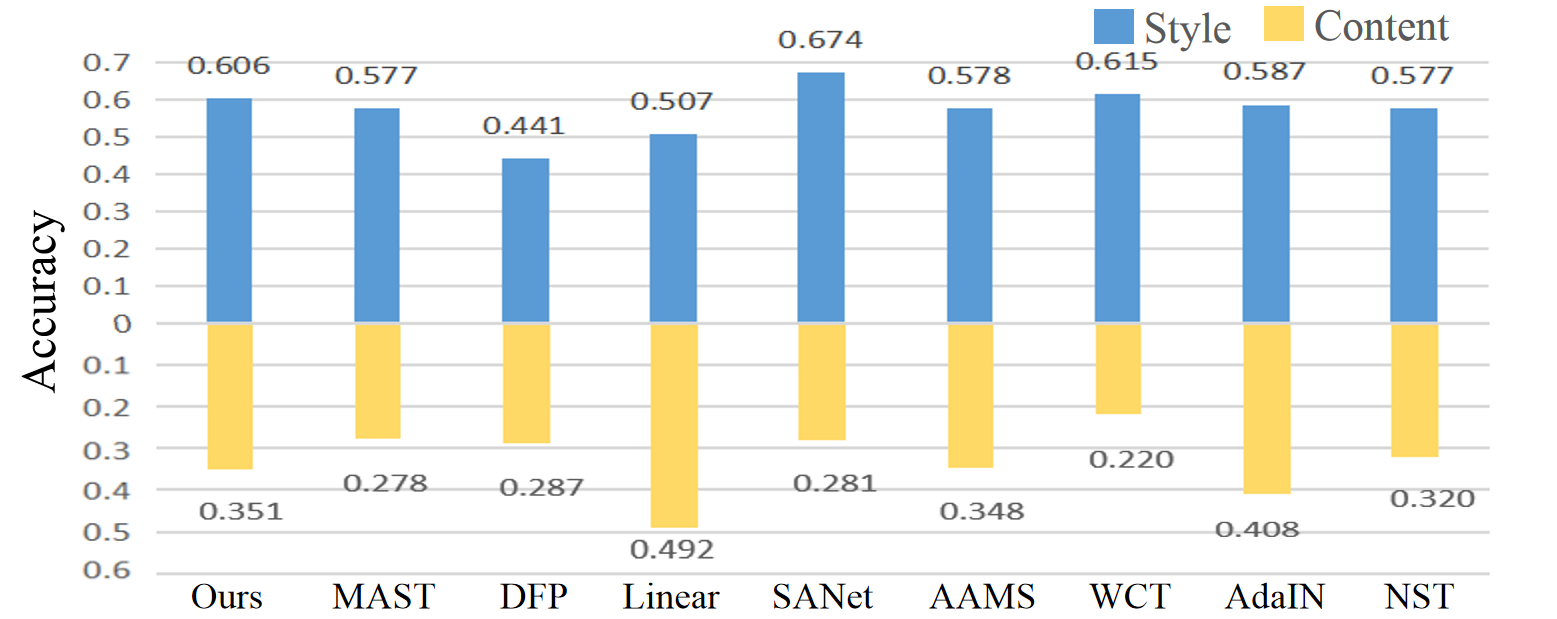}
\caption{Classification accuracy.}
\label{fig:class}
\end{figure}

\paragraph{Timing information.}
We measure our inference time for the generation of an output image and compare the result with those of SOTA methods using 16G TitanX GPU.
The optimization-based method NST~\citep{gatys:2016:image} is trained for $30$ epochs.
Table~\ref{tab:speed} shows the inference times of different methods using three scales of image size.
The inference speed of our network is much faster than that in \citep{gatys:2016:image, li:2017:universal, yao:2019:attention,wang:2020:diversified, wang:2020:consistent, deng:2020:arbitrary}.
Our method can achieve a real-time transfer speed that is comparable to that of \citep{Huang:2017:Arbitrary,li:2019:learning, park:2019:arbitrary} for efficient video style transfer.

\begin{figure}
\setlength{\abovecaptionskip}{2mm}
\centering
\includegraphics[width=\linewidth]{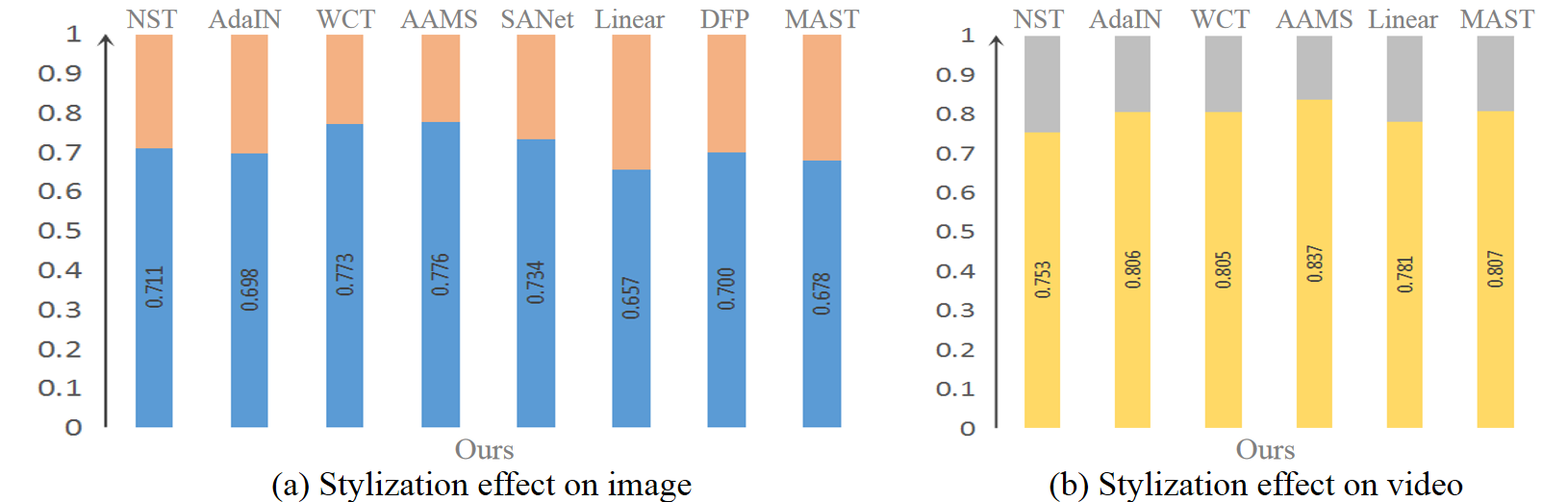}
\caption{User study results.}
\label{fig:user}
\end{figure}

\subsection{Image Style Transfer Results}

\paragraph{Qualitative analysis.}
CompoundVST is not selected for image stylization comparison because it is only used for video style transfer.
The comparisons of image stylizations are shown in Figure~\ref{fig:artist}.
On the basis of the optimized training mechanism, NST may introduce failure results (the second row), and it cannot easily achieve a trade-off between the content structure and style patterns in the rendered images.

In addition to crack effects caused by over-simplified calculation in AdaIN, the absence of correlation between different channels causes a poor transfer of style color distribution badly transferred (the 4th row).
As for WCT, the local content structures of generated results are damaged due to the global parameter transfer method.
Although the attention map in AAMS helps to make the main structure exact, the patch splicing trace affects the overall generated effect.
Some style image patches are transferred into the content image directly (the first row) by SANet and damage the content structures (the fourth row). 
By learning a transformation matrix for style transfer, the process of Linear is too simple to acquire adequate style textural patterns for rendering. 
DFP focuses on generating diversified style transfer results, but it may lead to failures similar to WCT.
MAST may introduce unexpected style patterns in rendered results (the fourth and the fifth rows).

Our network calculates the multi-channel correlation of content and style features and rearranges style features on the basis of the correlation.
The rearranged style features fit the original content features, and the fused generated features consist of clear content structures and controllable style pattern information.
Therefore, our method can achieve satisfactory stylized results.

\begin{figure}
\setlength{\abovecaptionskip}{2mm}
\centering
\includegraphics[width=\linewidth]{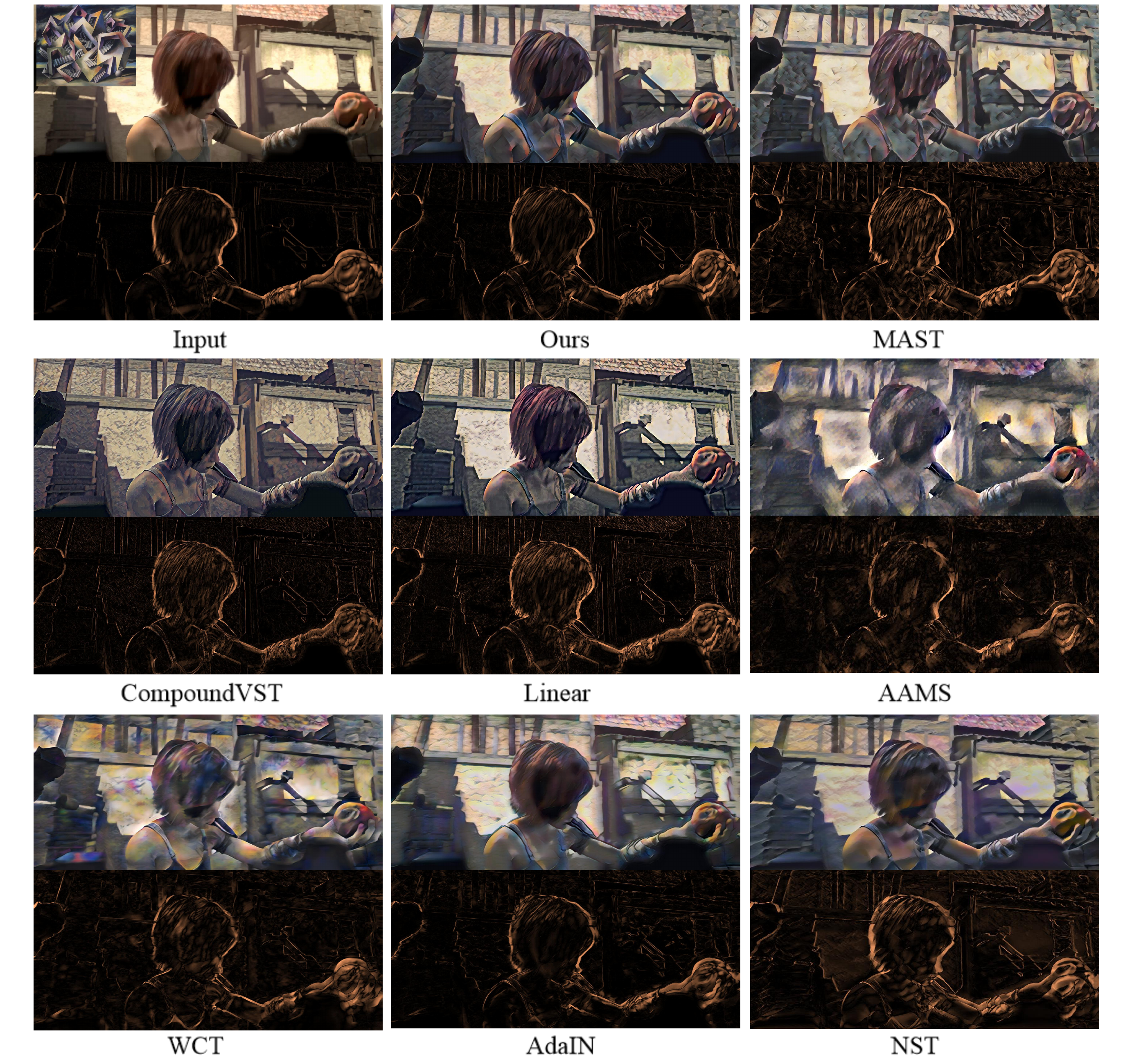}
\caption{Visual comparisons of video style transfer results. The first row shows the video frame stylized results. The second row shows the heat maps which are used to visualize the differences between two adjacent video frame.}
\label{fig:video}
\end{figure}

\paragraph{Quantitative analysis.}
Two classification models are trained to assess the performance of different style transfer networks by considering their ability to maintain content structure and style migration.
We generate several stylized images by using different style transfer methods aforementioned.
Then we input the stylized images generated by different methods. Then, we input the stylized images generated by different methods into the style and content classification models.
High-accuracy style classification indicates that the style transfer network can easily learn to obtain effective style information while high-accuracy content classification indicates that the style transfer network can easily learn to maintain the original content information.
From Figure~\ref{fig:class}, we can conclude that AAMS, AdaIN, and MCCNet can establish a balance between content and style. 
However, our network is superior to the two methods in terms of visual effects.

\paragraph{User study}
We conducted user studies to compare the stylization effect of our method with those of the aforementioned SOTA methods.
We selected $20$ content images and $20$ style images to generate $400$ stylized images using different methods.
First, we showed participants a pair of content and style images.
Second, we showed them two stylized results generated by our method and a random contrast method.
Finally, we asked the participants which stylized result has the best rendered effects by considering the integrity of the content structures and the visibility of the style patterns.
We collected $2,500$ votes from $50$ participants and present the voting results in Figure~\ref{fig:user}(a).
Overall, our method can achieve the best image stylization effect.

\begin{table*}
\small
\setlength\tabcolsep{2.5pt}
\centering
\begin{tabular}{c|c|cccccccc}
\toprule
&Inputs&Ours&MAST&CompoundVST& Linear& AAMS & WCT&AdaIN&NST \\
\midrule
Mean & 0.0143&\textbf{0.0297}&0.0548&0.0438& 0.0314& 0.0546&0.0498&0.0486&0.0417 \\
Variance & 0.0022&\textbf{0.0054}&0.0073&0.0054& 0.0061& 0.0067&0.0061&0.0067&0.0065 \\
\bottomrule
\end{tabular}
\caption{Average $mean_{Diff}$ and $var_{Diff}$ of inputs and $14$ rendered clips.}
\label{tab:difference}
\end{table*}

\subsection{Video Style Transfer Results}
Considering the size limitation for input images of SANet and generation diversity of DFP, we do not include SANet and DFP for video stylization comparisons.
We synthesize $14$ stylized video clips by using the other methods 
and measure the coherence of the rendered videos by calculating the differences in the adjacent frames.
As shown in Figure~\ref{fig:video}, the heat maps in the second row visualize the differences between two adjacent frames of the input or stylized videos.
Our method can highly promote the stability of image style transfer.
The differences of our results are closest to those of the input frames without damaging the stylization effect.
MAST, Linear, AAMS, WCT, AdaIN, and NST fail to retain the coherence of the input videos.
Linear can also generate a relatively coherent video, but the result continuity is influenced by nonlinear operation in deep CNNs.

Given two adjacent frames $F_{t}$ and $F_{t-1}$ in a T-frame rendered clip, we define ${Diff}_{F(t)}=||F_{t}-F_{t-1}||$ and calculate the mean ($mean_{Diff}$) and variance ($var_{Diff}$) of ${Diff}_{F(t)}$.
As shown in Table~\ref{tab:difference}, we can conclude that our method can yield the best video results with high consistency.

\paragraph{User study.}
The global feature sharing used in CompoundVST increases the complexity of the model and limits the length of video clips that can be processed. 
Therefore, CompoundVST is not selected for comparison in this section.
Then we conducted user studies to compare the video stylization effects of our method with those of the aforementioned SOTAs. 
First, we showed the participants an input video clip and a style image.
Second, we showed them two stylized video clips generated by our method and a random contrast method.
Finally, we asked the participants which stylized video clip is the most satisfying by considering the stylized effect and the stability of the videos.
We collected $700$ votes from $50$ participants and present the the voting results in Figure~\ref{fig:user}(b).
Overall, our method can achieve the best video stylization effect. 

\subsection{Ablation Study}
\label{sec:ablation}
\begin{figure}
\setlength{\abovecaptionskip}{2mm}
\centering
\includegraphics[width=\linewidth]{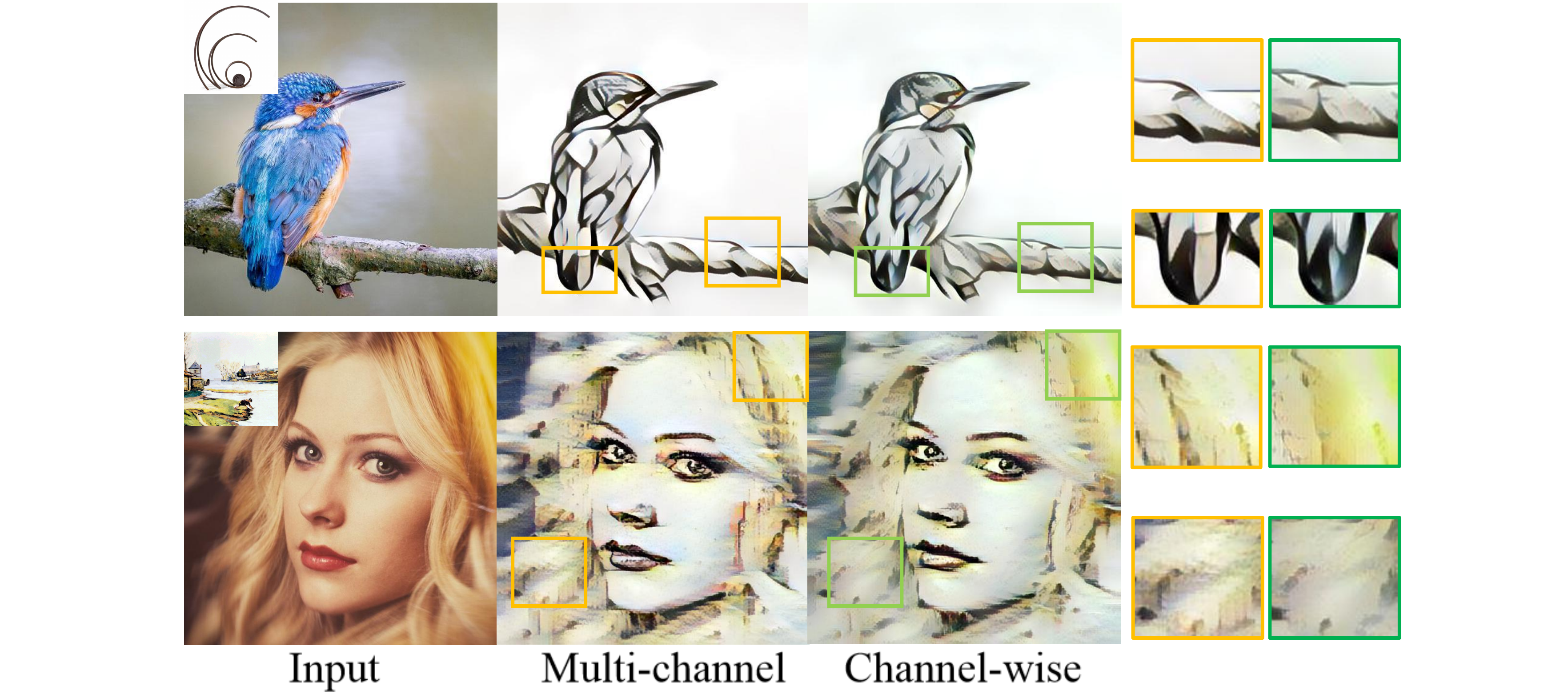}
\caption{Ablation study of channel correlation. 
Without multi-channel correlation, the stylized results present less style patterns (e.g., hair of the woman) and may maintain original color distribution (the blue color in the bird's tail).}
\label{fig:channel}
\end{figure}

\paragraph{Channel correlation.}
Our network is based on multi-channel correlation calculation shown in Figure~\ref{fig:technical}.
To analyze the influence of multi-channel correlation on stylization effects, we change the model to calculate channel-wise correlation without considering the relationship between style channels.
Figure~\ref{fig:channel} shows the results.
Through channel-wise calculation, the stylized results show few style patterns (e.g., hair of the woman) and may maintain the original color distribution (the blue color in the bird's tail). Meanwhile, the style patterns are effectively transferred by considering the multi-channel information in the style features.

\begin{figure}
\setlength{\abovecaptionskip}{2mm}
\centering
\includegraphics[width=\linewidth]{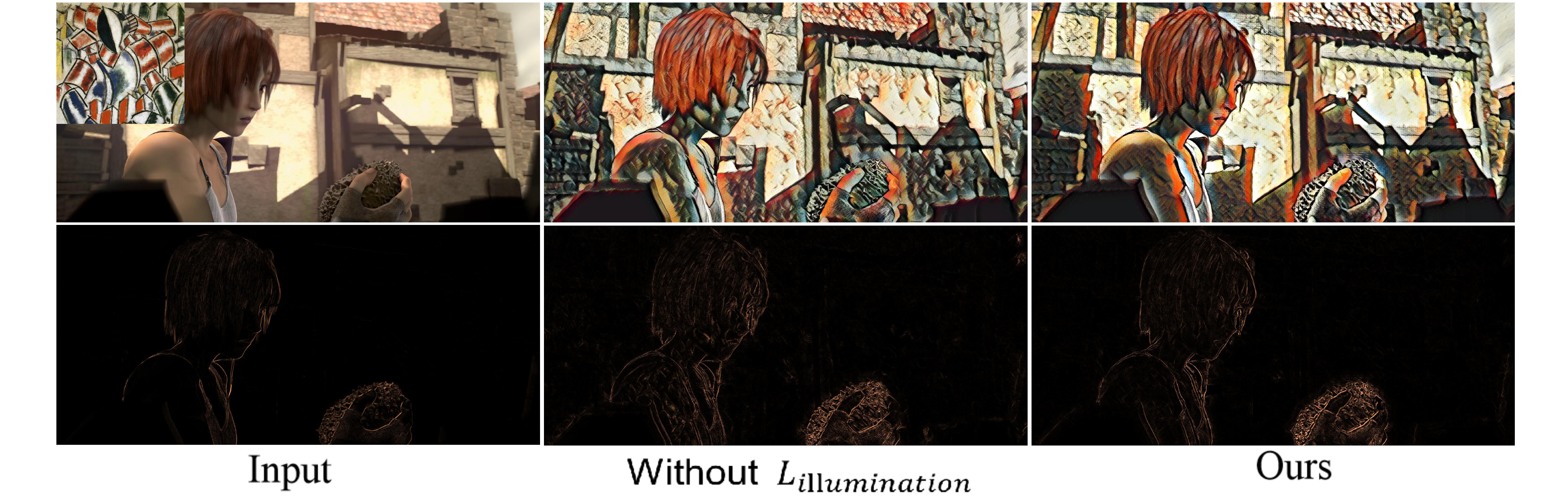}
\caption{Ablation study of removing the illumination loss. The bottom row is heat maps used to visualize the differences between the above two video frames.}
\label{fig:light}
\end{figure}

\paragraph{Illumination loss.}
The illumination loss is proposed to eliminate the impact of video illumination variation.
We remove the illumination loss in the training stage and compare the results with ours in Figure~\ref{fig:light}. 
Without illumination loss, the differences between the two video frames increase, with the mean value being $0.0317$.
With illumination loss, our method can be flexibly applied to videos with complex light conditions.

\begin{figure}
\setlength{\abovecaptionskip}{2mm}
\centering
\includegraphics[width=\linewidth]{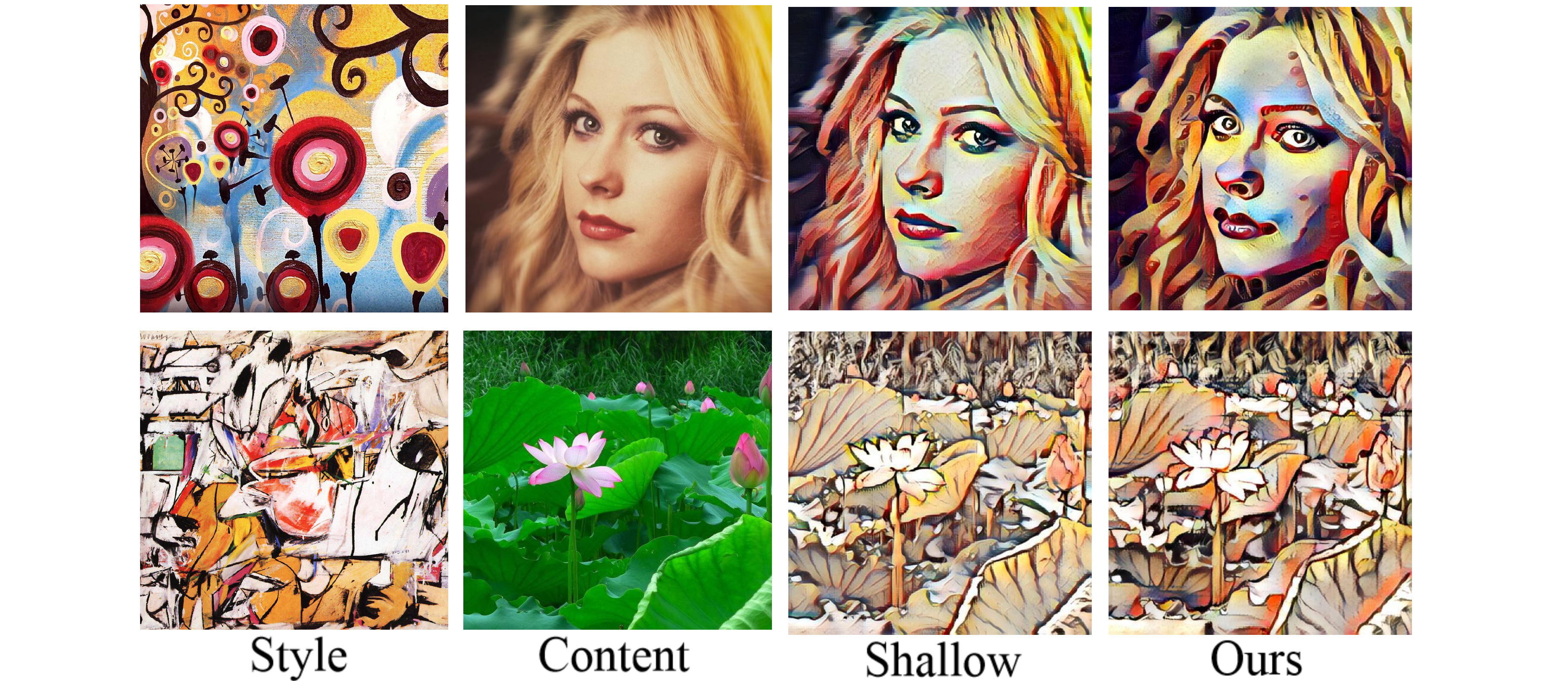}
\caption{Ablation study of network depth. Compared to the shallower network, our method generates artistic style transfer results with more stylized patterns.}
\vspace{-3mm}
\label{fig:shallower1}
\end{figure}

\begin{figure}
\setlength{\abovecaptionskip}{2mm}
\centering
\includegraphics[width=\linewidth]{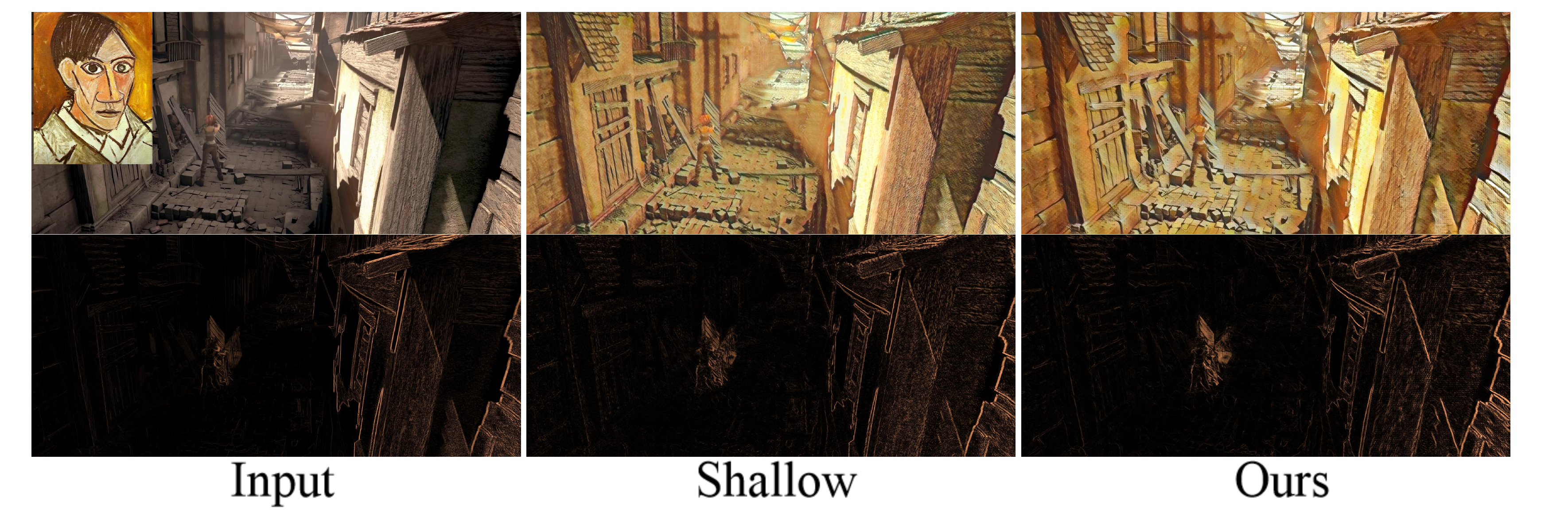}
\caption{Ablation study of network depth. The heat maps in the second row show the differences between two video frames in the first row.
}
\label{fig:shallower2}
\end{figure}

\paragraph{Network depth.}
We use a shallow auto-encoder up to $relu3$-$1$ instead of $relu4$-$1$ to discuss the effects of the convolution operation of the decoder on our model.
As shown in Figure~\ref{fig:shallower1}, for image style transfer, the shallower network can not generate results with vivid style patterns (e.g., the circle element in the first row of our results).
As shown in Figure~\ref{fig:shallower2}, the depth of the network exerts little impact on the coherence of the stylized video.
This phenomenon suggests that the coherence of stylized frames features can be well-transited to the generated video despite the convolution operation of the decoder.

\section{Conclusion}
In this work, we propose MCCNet for stable arbitrary video style transfer. 
The proposed network can migrate the coherence of input videos to stylized videos and thereby guarantee the stability of rendered videos. 
Meanwhile, MCCNet can generate stylized results with vivid style patterns and detailed content structures by analyzing the multi-channel correlation between content and style features.
Moreover, the illumination loss improves the stability of generated video under complex light conditions.

\section*{Acknowledgements}
This work was supported by National Key R\&D Program of China under no. 2020AAA0106200, and by National Natural Science Foundation of China under nos. 61832016, U20B2070 and 61672520.

\bibliography{references}

\end{document}